\documentclass[a4paper]{llncs}
\usepackage{bibnames}
\usepackage{amssymb}
\usepackage{graphicx}
\usepackage{url}
\usepackage{xcolor}
\urldef{\mailsa}\path|{alfred.hofmann, ursula.barth, ingrid.haas, frank.holzwarth,|
\urldef{\mailsb}\path|anna.kramer, leonie.kunz, christine.reiss, nicole.sator,|
\urldef{\mailsc}\path|erika.siebert-cole, peter.strasser, lncs}@springer.com|    
\newcommand{\keywords}[1]{\par\addvspace\baselineskip
\noindent\keywordname\enspace\ignorespaces#1}

\usepackage{cite}
\usepackage{amsmath,amssymb,amsfonts}
\usepackage{graphicx}
\usepackage{textcomp}
\usepackage{xcolor}
\usepackage[textwidth=3.1em]{todonotes}
\usepackage{tikz}
\usepackage{subfiles}
\usepackage{pgfplotstable}
\usepackage{subcaption}
\usepackage{amsmath,amssymb,amsfonts}
\usepackage{algpseudocode} 
\usepackage{siunitx,booktabs} 
\usepackage{setspace} 
\usepackage{color,soul} 
\usepackage{flushend}
\usepackage{tikz,pgfplots}
\usepackage{pdfpages}
\usepackage{lineno}

\usepackage{pgfplots}
\usetikzlibrary{patterns}
\pgfplotsset{width=6cm, compat=1.6}

\begin{document}

\mainmatter  

\title{Edge-centric Optimization of Multi-modal ML-driven eHealth Applications}


%

%
\author{Anil Kanduri$^{\ddag}$, Sina Shahhosseini$^{\dag}$, Emad Kasaeyan Naeini$^{\dag}$, Hamidreza Alikhani$^{\dag}$, Pasi Liljeberg$^{\ddag}$, Nikil Dutt$^{\dag}$, and Amir M. Rahmani$^{\dag,\S}$}
\authorrunning{Anil Kanduri, Sina Shahhosseini, Emad Kasaeyan Naeini, Hamidreza Alikhani, Pasi Liljeberg, Nikil Dutt, and Amir M. Rahmani}

\institute{
$^{\dag}$ Department of Computer Science, University of California, Irvine\\
Mailing address: Bren Hall 3086, Irvine, CA 92697-3435, USA\\
$^{\S}$ School of Nursing, University of California, Irvine\\
Mailing address: 106D Berk Hall, Irvine, CA 92697, USA\\
$^{\ddag}$ Department of Computing, University of Turku, Finland\\
Mailing address: Vesilinnantie 3, Turku 20540, Finland\\ 
Email: spakan@utu.fi, sshahos@uci.edu, ekasaeya@uci.edu, hamidra@uci.edu, pasi.liljeberg@utu.fi, dutt@uci.edu, a.rahmani@uci.edu
}


\maketitle

\begin{abstract}
Smart eHealth applications deliver personalized and preventive digital healthcare services to clients through remote sensing, continuous monitoring, and data analytics. 
Smart eHealth applications sense input data from multiple
modalities, transmit the data to edge and/or cloud nodes, and process the data with compute intensive machine learning (ML) algorithms. Run-time variations with continuous stream of noisy input data, unreliable network connection, computational requirements of ML algorithms, and choice of compute placement among sensor-edge-cloud layers affect the efficiency of ML-driven eHealth applications. In this chapter, we present edge-centric techniques for optimized compute placement, exploration of accuracy-performance trade-offs, and cross-layered sense-compute co-optimization for ML-driven eHealth applications. We demonstrate the practical use cases of smart eHealth applications in everyday settings, through a sensor-edge-cloud framework for an objective pain assessment case study.

\end{abstract}

\section{Introduction}

Smart eHealth applications deliver critical digital healthcare services such as disease diagnostics, clinical decision support, forecasting health status, pro-active and preventive healthcare decisions, and alerts for emergency intervention, etc \cite{farahani2020towards}. eHealth applications improve the reach and quality of healthcare services, timeliness and accuracy of clinicians decisions, and reduce the burden on healthcare professionals and overall medical expenditure \cite{versluis2020series}. Smart eHealth systems integrate remote sensing, continuous monitoring, wireless transmission, data analytics, and machine learning to deliver intelligent patient-centric digital healthcare and wellbeing services \cite{khan2021rebirth}. eHealth applications are particularly effective for managing chronic patients through continuous monitoring, extracting clinically relevant data with minimal intrusion \cite{duch2017heal}.

\subsection{ML in Smart eHealth Applications}
\begin{figure}[t]
    \centering
    \includegraphics[width=0.75\linewidth]{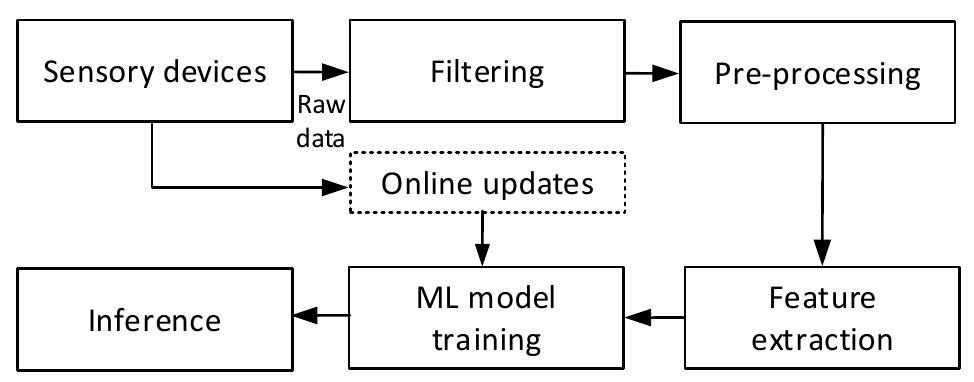}
    \caption{ML-driven eHealth application pipeline}
    \label{fig.ml-pipeline}
\end{figure}

eHealth systems continuously monitor patients using wearable sensors for acquiring physiological parameters \cite{dogan2012multi}. In addition to the bio-signals, eHealth applications also track behavioral, and environmental parameters to contextualize the patients' current situation \cite{kreps2010new}. Thus, smart eHealth applications generate huge volumes of heterogeneous input data, combining multiple streams of inputs from physiological, behavioral, and environmental parameters \cite{farahani2020towards}. Analyzing such continuous stream of heterogeneous multi-modal raw data for predicting potential threats, accurate clinical decisions, and diagnostics requires broader support from the AI domain \cite{gupta2022artificial}. Smart eHealth systems are increasingly using ML algorithms for analyzing multi-modal input sensory data, to provide intelligent digital healthcare and wellbeing services \cite{khan2021rebirth}. State-of-the-art eHealth applications have applied different ML algorithms for analyzing input data, and predicting results on diagnostics, potential and health status \cite{chetty2015intelligent}. ML-driven eHealth systems work in a pipeline of data acquisition, filtering and pre-processing, data analysis, training, inference for predictive results, followed by notification to the clients \cite{farahani2020towards}. Figure \ref{fig.ml-pipeline} shows the workflow of typical ML-driven eHealth applications, where raw data acquired by sensory devices is filtered and pre-processed to remove noisy components, motion artifacts, and anomalies. This input data is then used for extracting relevant features, and training suitable ML models. Predictive results are achieved by inferencing the trained ML model, while the trained model is periodically updated with evolving input data. 

\subsection{Collaborative Edge Computing for Smart eHealth Applications}
\begin{figure}[t]
    \centering
    \includegraphics[width=0.98\linewidth]{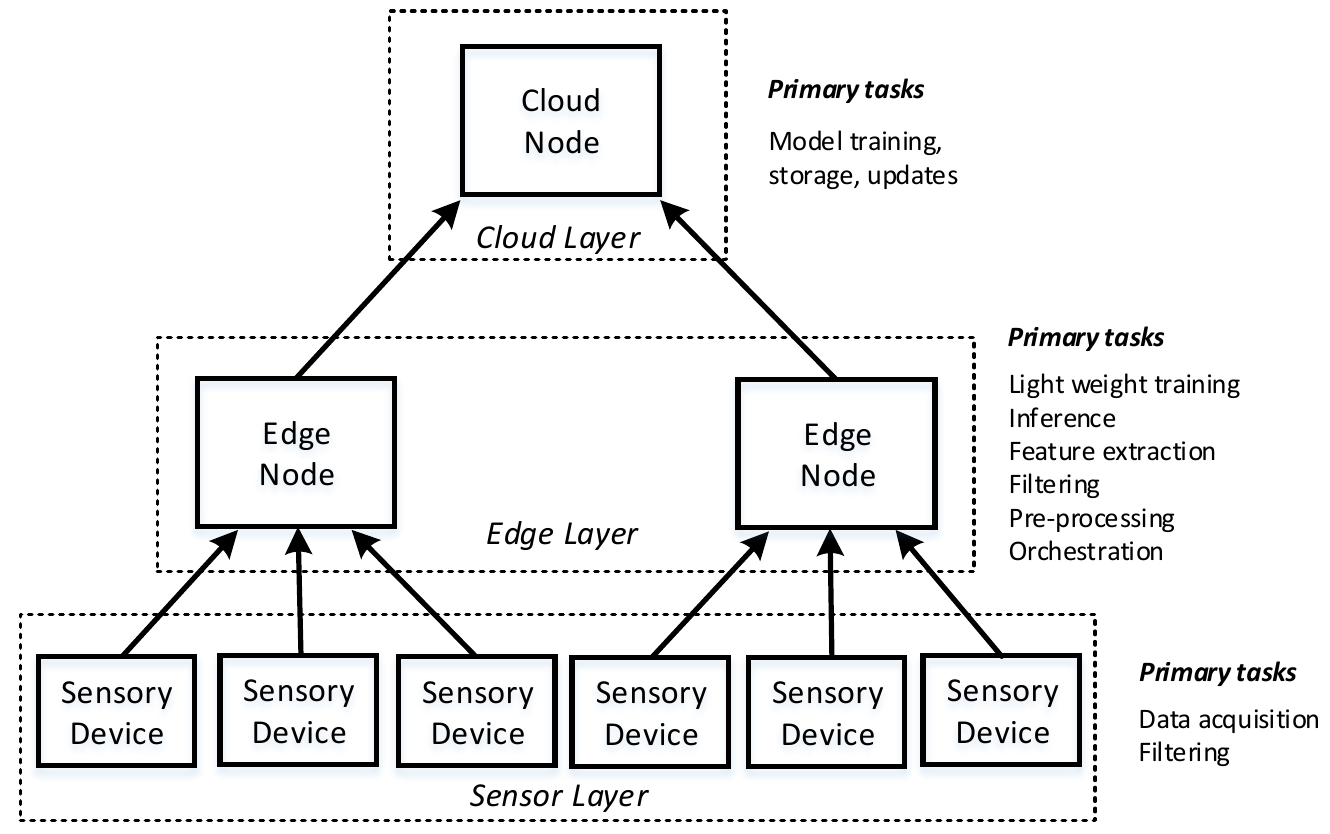}
    \caption{Sensor-edge-cloud architecture}
    \label{fig.3-layer-arch}
\end{figure}

eHealth applications rely on traditional cloud infrastructure for training, storage, and updating of ML models, and inference tasks. However, rapidly increasing volumes of sensory input data, and uncertain network conditions imposes limitations on the efficacy of running eHealth applications on the cloud layer. Edge computing paradigm brings computational intelligence closer to the sources of input data, minimizing the reliance of smart eHealth applications on cloud infrastructure \cite{wang2020convergence}. Edge computing architectures have been widely adopted for deploying ML-driven smart eHealth applications, simultaneously handling input data, compute intensity, and network constraints \cite{rahmani2018exploiting}. Figure \ref{fig.3-layer-arch} shows an overview of the hierarchical multi-layered sensor-edge-cloud architecture \cite{azimi2017}. The \textit{sensor layer} comprises of sensory devices such as wearable sensors, smart bio-sensors, and sensors deployed on mobile and IoT devices. The \textit{sensor layer} primarily acquires raw data from different devices, performs lightweight tasks such as filtering, and transmits relevant inputs to the resourceful layers in the hierarchy. The \textit{edge layer} receives input data from the sensory devices, and executes intensive tasks such as data pre-processing, feature extraction, lightweight ML model training, and inference tasks. More importantly, the \textit{edge layer} also handles orchestration functionalities such as application-level and system-level monitoring, application partitioning, compute placement, and resource allocation. The \textit{cloud layer} handles heavy computational tasks such as ML model training, updating, and storage, and notifications to edge nodes on model updates.


%




\subsubsection{Example Scenario:}


\begin{figure}
    \centering
    \includegraphics[width=0.8\linewidth]{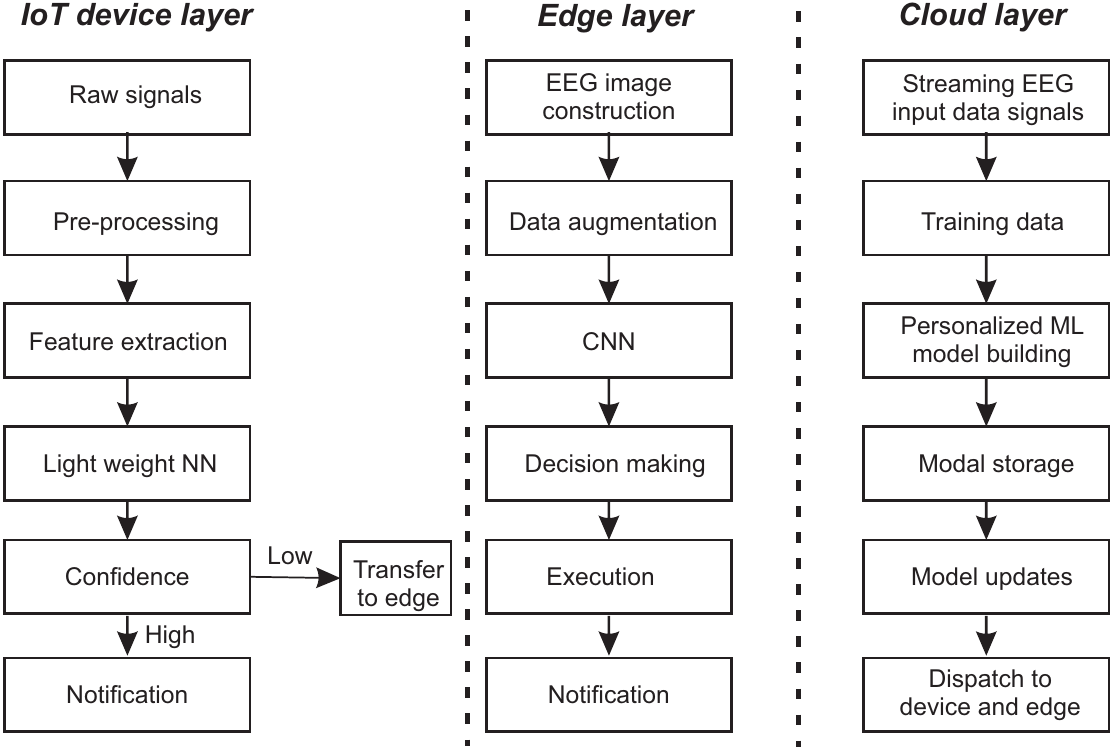}
    \caption{Pipeline of arrhythmia detection}
    \label{fig.aryth_example}
\end{figure}

We demonstrate the pipeline of a quintessential edge-centric ML-driven eHealth application through the example of arrhythmia detection application \cite{farahani2020towards}. Figure \ref{fig.aryth_example} shows the data and control flows of the arrhythmia detection application \cite{farahani2020towards} across the device-edge-cloud layers. Initially, raw electroencephalogram (EEG) signals are acquired by wearable sensory devices. The raw signals are pre-processed for noise removal to extract relevant features from the raw data for training an ML model. Considering the limited compute resources of the IoT device layer, a lightweight neural network (NN) model is employed to predict/detect arrhythmia. The predictions are notified to the client if the NN model has a higher confidence on prediction accuracy, while forwarding the input data to edge layer when the model confidence is lower. In this application, the edge layer uses the reconstructed EEG images for data aggregation. With relatively higher compute resources, the edge layer consists of a moderately intensive convolutional neural network (CNN) model to train on input data. The CNN model is inferred in the execution phase for arrhythmia detection/prediction, and the client is notified with the result. The cloud layer collects streaming inputs from the device and edge layers to train appropriate ML model, store the model, and update the model periodically with evolving input data. The cloud layer transmits the updated model parameters to the edge and device layers for running local inference tasks. Further the cloud layer performs advanced data analytics to generate personalized decisions for each client. It should be noted that the compute capabilities and thus tasks vary at each of the device-edge-cloud layers. Optimizing ML-driven eHealth systems requires such understanding on input data flows, computational requirements, and accuracy and performance of ML models.

\subsection{Summary}
Implementing ML-driven smart eHealth applications presents different challenges on sensory data acquisition, understanding application-level requirements, handling compute intensities of ML models, and energy and network constraints. At the same time, deploying such applications on multi-layered sensor-edge-cloud platforms exposes opportunities for selective processing through input data quality awareness, choice of compute placement among edge-cloud nodes exploring energy-performance trade-offs, and understanding algorithmic nature of applications to explore accuracy-performance-energy trade-offs. Collaborative sensor-edge-cloud platforms enable layer-wise partitioning of smart eHealth application pipeline, to synergistically improve the quality of the services. In the subsequent sections, we present different edge-centric optimizations for ML-driven smart eHealth applications on collaborative sensor-edge-cloud platforms.
\subsubsection{Organization:} Section \ref{sec.case-study} presents an exemplar case study of pain assessment application, describing a sensor-edge-cloud framework integrating different edge-centric optimizations. This case study is then used in the following chapters to demonstrate some of the optimization techniques. Section \ref{sec.edge} presents techniques for improving performance metrics of edge-centric ML workloads through efficient compute placement, and exploration of accuracy-latency trade-offs. Section \ref{sec.amser} presents techniques for improving resilience of ML-driven eHealth applications, through sense-compute co-optimization. Section \ref{sec.conclusion} concludes with key insights, and open research directions. 

\section{Exemplar Case Study of Edge-ML-driven Pain Assessment Application} \label{sec.case-study}
In this section, we present an exemplar case study of pain assessment application, describing application characteristics, nature of input data, and challenges in typical edge-centric ML-driven smart eHealth application. We also present a modular framework, iHurt, for deploying ML-driven eHealth applications (pain assessment in this case study) on collaborative sensor-edge-cloud platforms. The iHurt framework serves as a generic platform for prototyping smart eHealth applications for processing sensory data using ML models. Further, iHurt platform provides a test-bed for evaluating the edge orchestration, compute placement, RL-agent based offloading, and sense-compute co-optimization techniques presented in Sections \ref{sec.edge} and \ref{sec.amser}.

\begin{figure}
    \centering
    \includegraphics[width=0.8\linewidth]{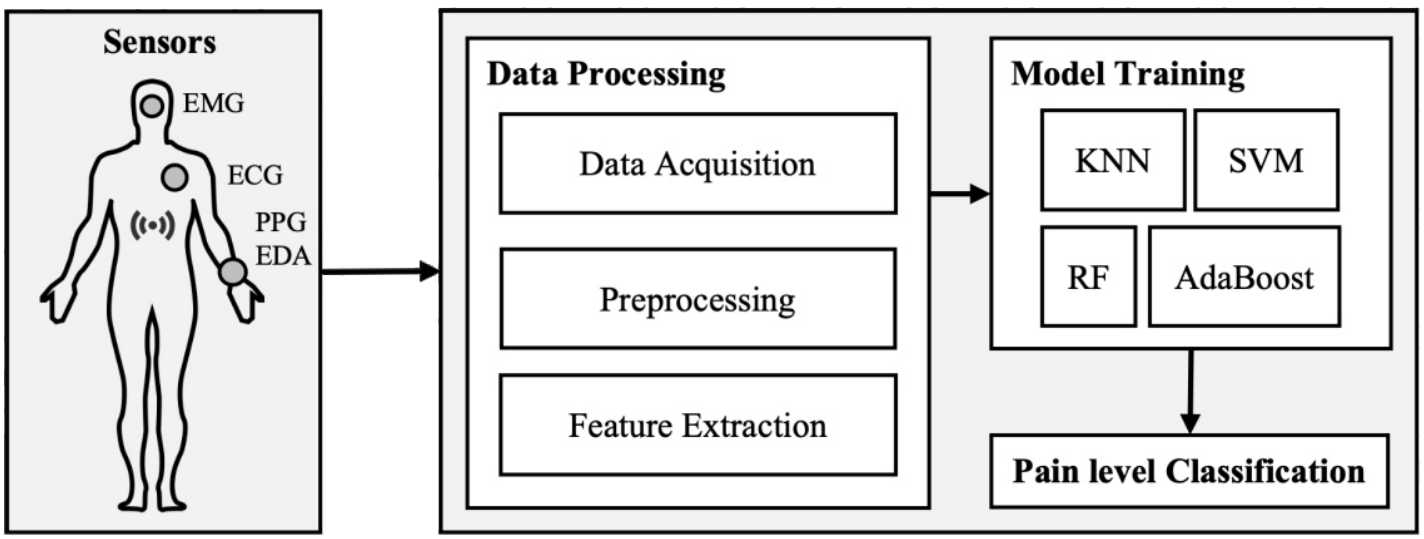}
    \caption{Overview of pain assessment application}
    \label{fig.pain_top}
\end{figure}

\subsection{Pain Assessment}
Pain is a complex phenomenon, associated with several illnesses \cite{TOMPKINS2017S11}. Pain is defined as "an unpleasant sensory and emotional experience associated with actual or potential tissue damage, or described in terms of such damage" \cite{merskey1979pain}. The pain assessment ``gold standard" relies on a patient's self-report of their pain intensity on a scale of 0 to 10, where 0 refers to no pain, and 10 represents the most severe pain. Pain assessment is done through tools such as Numerical Rating Scale (NRS), Visual Analogue Scale (VAS), and Verbal Rating Scale (VRS). There is a high demand for objective tools to assess patients’ pain in the clinical context. Tools are needed especially when the patient’s own opinion is difficult to obtain. Assessment of pain is particularly difficult when the ability of a patient to communicate is limited (e.g., during critical illness, in infants and preverbal toddlers, or in patients under sedation or anesthesia, with intellectual disabilities, and at the end of life) \cite{breivik2008assessment}. Inadequately treated pain has major physiological, psychological, economic, and social ramifications for patients, their families, and society \cite{arif2010facial}. 
Undertreatment of pain could result in many adverse effects and other complications and may evolve into chronic pain syndromes. It could also cause delayed discharge or prolonged recovery, which may incur higher health care costs and more patient suffering \cite{stites2013observational}. Overtreatment of pain, on the other hand, may result in unintended adverse consequences such as acute respiratory complications or in long-term complications such as opioid addiction. These issues are particularly pronounced for noncommunicative patients who are unable to articulate their experience of pain \cite{barr2013clinical}. 

We demonstrate an abstract overview of the pain assessment application \cite{jrp} in Figure \ref{fig.pain_top}. The pain assessment application is implemented as a pipeline of sensing, data processing, model training for predictive results, and inference for pain level classification. In the context of this case study, we use input data from EMG, ECG, PPG, and EDA sensors for estimation of pain. Sensory data is pre-processed for filtering qualitative inputs, followed by feature extraction. Relevant learning models are trained with the multi-modal input data sets for classifying pain level. Different phases of the pain assessment application are detailed in the following.

\subsection{Sensory Data Acquisition} Objective pain assessment application collects raw data through continuous monitoring in both clinical and every day settings \cite{farahani2020towards}. Raw data collected from the sensory data acquisition phase is used for training the ML models for accurate prediction of pain level. In this sub-section, we describe the nature and characteristics of sensory input data used in the pain assessment application.  


\subsubsection{Types of Signals:}



There are various types of signals that influence the accuracy of monitoring and assessing the affective states in pain assessment. These indicators are extracted through different forms of behavioral, physiological, and contextual/environmental sensor modalities via facial expression, speech, full-body motion, text, and physiological signals. Both behavioral and physiological manifestations of pain can be measured objectively. Behavioral pain indicators include facial expressions, body movements such as rubbing, restlessness, and head movements, and paralinguistic vocalizations such as crying and moaning. Physiological pain indicators are acquired from brain,  cardiovascular, and electrodermal activities. Monitoring of these physiological, behavioral, and contextual sensor inputs can also be used other prominent eHealth applications including emotion recognition and stress monitoring. For instance, stress activates the autonomic nervous system (ANS) which can be detected through monitoring the changes in physiological signals including cardiovascular activity, and electrodermal activity, respiration rate, and blood pressure \cite{greene2016survey}. Further, physiological signals of cardiac function, temperature, muscle electrical activity, respiration, skin conductance, and brain electrical activity can be used to detect human emotions. The multitude of physiological, behavioral, and contextual signals fused together can provide valuable insights for training ML models, specifically in the domain of smart affective computing applications.  

\subsubsection{Commonly Used Sensors:}

\begin{figure}[t]
    \centering
    \includegraphics[width=0.8\linewidth]{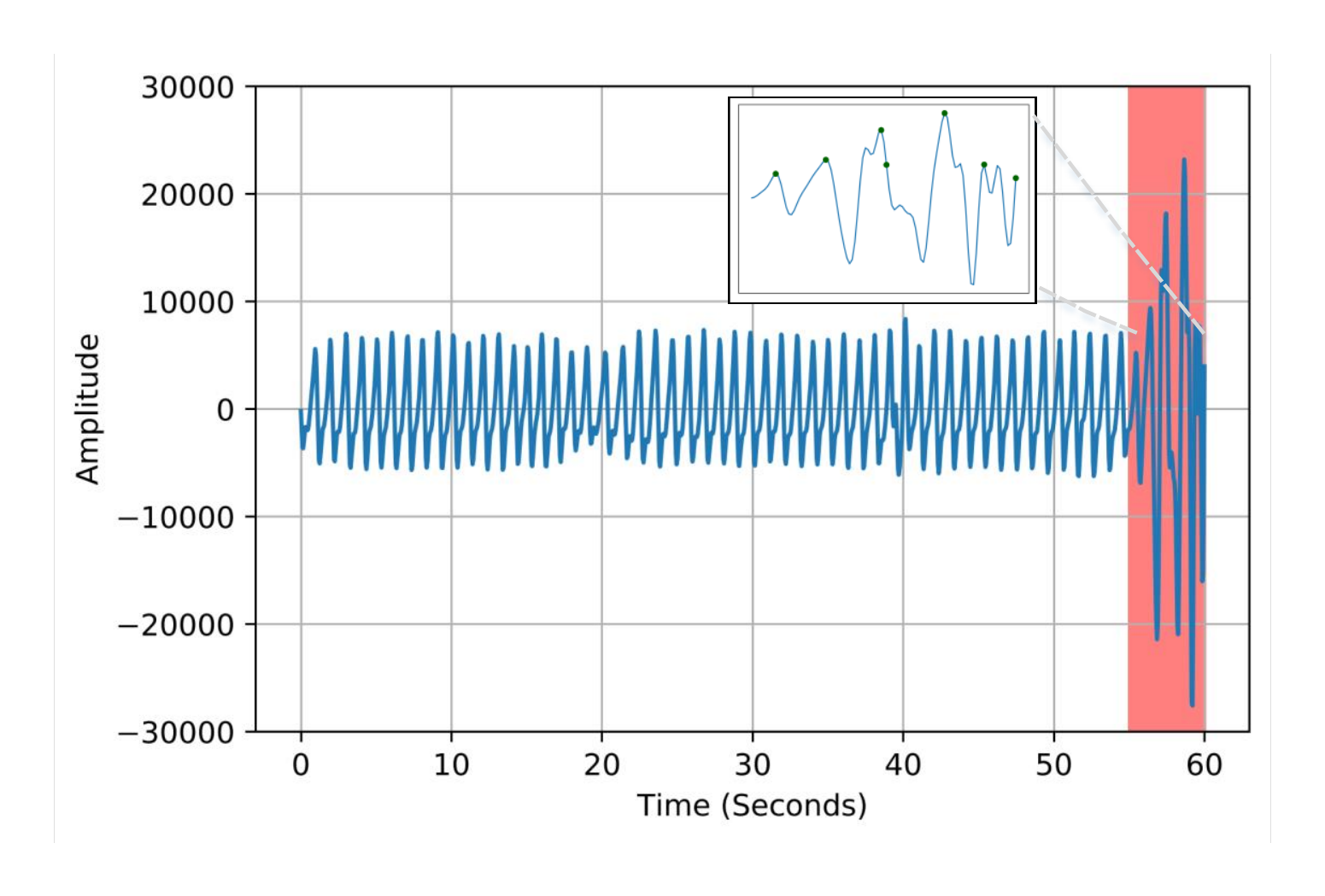}
    \caption{One-minute windows of filtered PPG signals carrying noise}
    \label{fig.ppg}
\end{figure}

Recording physiological signals require people to connect with biosensors. There are contact-based sensors (such as adhesive electrodes and wristbands) or contact-free sensors (such as cameras and microphones) to gather information from patients and analyze them. Widely used contact sensors in eHealth applications record electroencephalogram (EEG, electrical activity of the brain), electrocardiogram (ECG, heart activity (heart rate (HR) and heart rate variability (HRV))), electrodermal activity (EDA often measured using skin conductance level (SCL), and sometimes the old term “galvanic skin response” (GSR)), surface electromyogram (sEMG, muscle activity), photoplethysmogram (PPG, blood perfusion of the skin for pulse and other measures, also called blood volume pulse or BVP), respiration (RSP), or acceleration (ACC, movement). For pain monitoring, the uni-dimensional assessment tools have been questioned and debated for their oversimplification and limited applicability in non-communicative patients, since they require interactive communication between patient and caregiver \cite{jiang2017ultra}. As a result, physiological sources of data comprising of heart rate (HR), heart rate variability (HRV), SpO2, skin temperature, electrodermal activity (EDA), and facial expression and frontal muscle activity using computer vision or facial electromyography (EMG) and electroencephalogram (EEG) are prioritized for pain assessment. Accurate calculation of HRV parameters depends on detecting the position of peaks within ECG or PPG signals. Root Mean Square of Successive Differences (RMSSD) is an HRV parameter that is correlated to the short-term variation in the PPG signal. Figure \ref{fig.ppg} shows a minute filtered PPG signal illustrating an example, in which less than 5 seconds of the PPG signal (highlighted in red) are distorted due to hand movements. Such a minor window of corrupted input data in the signal could still affect the eventual accuracy significantly. For instance, in Figure \ref{fig.ppg}, few peaks are detected incorrectly within the noisy signal part, and thus the RMSSD is not reliable anymore during this window of data. The pain assessment application uses ML models to detect such abnormalities, and enable accurately predictions.

\begin{table}[h!]
\centering
\resizebox{0.45\textwidth}{!}{\begin{tabular}{|c|c|c|c|c|}
\hline
            & \textbf{EDA} & \textbf{ECG} & \textbf{RR} & \textbf{MM}\\
            \hline
\textbf{BL vs. PL1} & 63.36 &{72.04} & {71.79} & {77.13}\\
\textbf{BL vs. PL2} & 79.24 & {81.13} & {82.14} & {85.64}\\
\textbf{BL vs. PL3} & {69.59} & {69.8} & {76.64} & {86.9}\\
\textbf{BL vs. PL4} & {63.7} & {63.41} & {66.67} & {74.73}\\
\hline
\textbf{Mean} & {68.97} & {71.6} & {74.31} & {81.1}\\
\hline
\end{tabular}}
\vspace{5pt}
\caption{Summary of accuracy results (BL: baseline, PL: pain level, MM: multi-modal)}
\label{tbl.pain}
\end{table}
\subsubsection{Multi-modal Inputs:} Objective pain assessment relies on input data from multiple modalities and combinations of physiological, behavioral, and contextual parameters. All these modalities differ in terms of data, noise characteristics, comfort and ease of use, privacy concerns, and energy consumption. Using a single modality versus a combination of multiple modalities effects the computational workloads of the ML models, and eventual prediction accuracy. This is demonstrated from a pain monitoring case study on a five levels of pain data collection \cite{naeini2020jrp,jmir_ecg,jmir_eda,embc_RRpain}. The accuracy of using each individual input modality of sensory data (EDA, ECG, RR), and multi-modal (MM) inputs for binary classification between no pain/baseline and various pain levels is shown in Table \ref{tbl.pain}. It should be noted that different sensor modalities result in a range of prediction accuracies across different levels of pain. Predictions based on multi-modal input data set has the highest accuracy among each of these cases. 

\subsection{ML-driven Objective Pain Assessment}
It is imperative to design and develop an objective monitoring tool to improve the well-being and care processes of patients with a more accurate assessment and more timely treatment. While this raises significant technical challenges, requiring a combination of sensing, signal processing, and machine-learning skill sets, it also has a tremendous potential to pave the way for next-generation human-modeling methods. Machine learning and deep learning techniques have become immensely popular for classification tasks, as well as for other recognition and pattern matching tasks. ML models can be used for accurate objective pain assessment, using input data from different modalities. The combination of vast amount of multi-modal input data, increased computing power and more intelligent methods enables fast and automated production of machine-learning algorithms able to analyze complex data with accurate results. 

\begin{figure}[ht!]
\centering
  \includegraphics[width=0.93\linewidth]{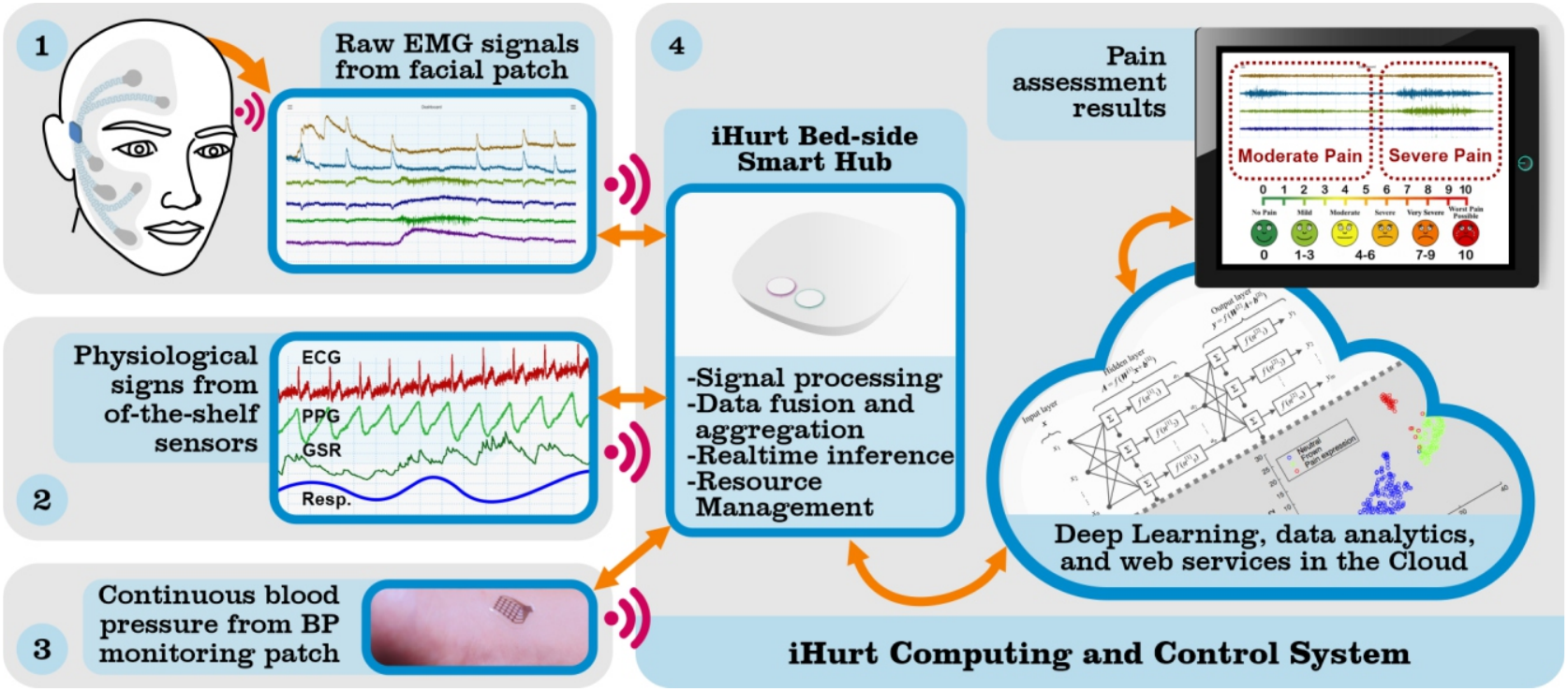}
  \caption{The Objective Pain Assessment Technology}
  \label{fig.ihurt_sys}
\end{figure}

\subsubsection{iHurt Platform}
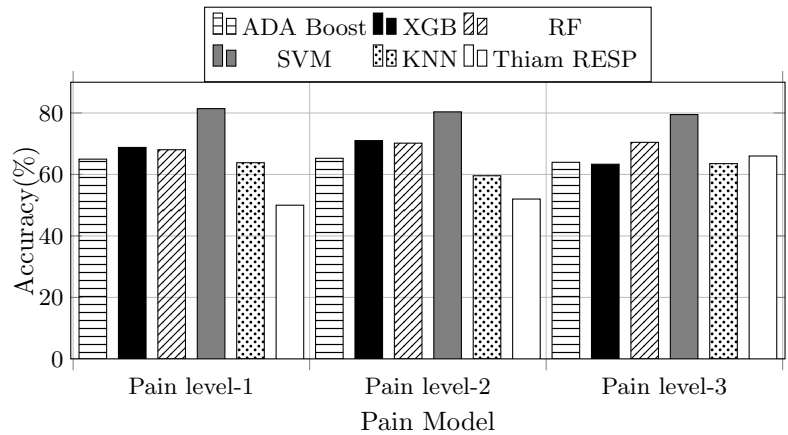
\begin{figure}[ht!]
\centering
\pgfplotsset{every x tick label/.append style={font=\tiny, yshift=0.2ex}}
\pgfplotsset{every y tick label/.append style={font=\tiny, xshift=0.2ex}}
\pgfplotsset{compat=1.3,
	/pgfplots/ybar legend/.style={
		/pgfplots/legend image code/.code={%
			\draw[##1,/tikz/.cd,yshift=-1mm]
			(5mm,2mm) rectangle (2pt,0.2em);},
	},
}
\begin{tikzpicture}
\begin{axis}[
enlarge x limits={abs=0.01},
width  = .9\textwidth,
height = 5.25cm,
bar width=5pt,
grid=major,
y label style={at={(-0.03,.5)},anchor=south},
xlabel={Pain Model}, ylabel={\normalsize Accuracy(\%)},
yticklabel style={rotate=0, font=\small},
xticklabel style={rotate=0, font=\small},
xtick = {0,1,2,3},
xticklabels={Pain level-1, Pain level-2, Pain level-3},
every node near coord/.append style={font=\tiny},   
	legend style={at={(0.5,1)},
	anchor = south, legend columns=3, font=\small},
	ybar interval=.7,
    ymin=0,ymax=90,
]
\addplot[draw=black,pattern=horizontal lines] 
	coordinates {(0,64.98)(1,65.26)(2,63.95)(3, 0)};
\addplot[draw=black,fill=black] 
	coordinates {(0,68.79)(1,71.01)(2,63.33)(3, 0)};
\addplot[draw=black,pattern=north east lines] 
	coordinates {(0,68.04)(1,70.18)(2,70.45)(3, 0)};
\addplot[draw=black,fill=gray] 
	coordinates {(0,81.41)(1,80.36)(2,79.48)(3, 0)};
\addplot[draw=black, pattern=crosshatch dots] 
	coordinates {(0,63.82)(1,59.58)(2,63.52)(3, 0)};
\addplot[draw=black,fill=white] 
	coordinates {(0,50)(1,52)(2,66)(3, 0)};
\legend{ADA Boost,XGB,RF,SVM,KNN,Thiam RESP}
\end{axis}
\end{tikzpicture}
\caption{Validation accuracy of classifiers on top-8 features for different pain levels compared against the baseline}
\label{fig:accuracy}
\end{figure}

Figure \ref{fig.ihurt_sys} presents a complete objective pain assessment technology developed jointly by researchers from University of California, Irvine (UCI) and University of Turku (UTU), Finland. This is an end-to-end system for multi-modal data acquisition, data processing and analyzing at the gateway. The first step in building a multimodal system is to process the raw signals collected during trials. While this varies according to the deployed sensor data, a typical pre-processing pipeline consists of the following: Filtering ( data cleansing, noise reduction, and artifact removal), segmentation (partitioning into time intervals) and normalization (w.r.t. to a baseline signal). These steps are followed by feature extraction to obtain features within various domains. Finally, the processed data can be used to build prediction models using machine learning. Different types of ML models can be used to train over multi-modal input data sets. Choice of the learning model determines the integration and fusion of multi-modal data at feature, decision, or intermediate levels (early, late, hybrid fusion). Most approaches classify pain using support vector machines (SVM) \cite{gruss2015pain, werner2013towards}, random forests (RF) \cite{werner2016automatic, kachele2015bio}, and nearest neighbors (NN) \cite{adibuzzaman2015assessment}. Other widely used models include Adaboost and XGBoost \cite{schapire2013explaining}, which are ensemble methods to reduce bias and variance in predictive data analysis. The ML models can be computationally fine-tuned differently to pursue various objectives. Sense-making knobs such as early exit, model selection, and input modality selection presented in Section \ref{sec.amser} can be explored in this context. Figure \ref{fig:accuracy} shows five different classification methods using respiratory signals including ADABoost, XGBoost, RF, SVM, and K-NN classifiers in comparison with a state-of-the-art method, RESP \cite{embc_RRpain}. This case study down sampled pain levels into three levels of pain (pain levels 1-3), besides the baseline of no pain. Note that they achieved higher accuracy compared with the state-of-the-art while using only 88\% less features. We can maintain accuracy in presence of noise, or using useful features from the reliable modalities, while also meeting the requirements set on the computational performance. The orchestration functionalities presented in Section \ref{sec.edge} can guide these decisions on exploring accuracy-performance trade-offs.




\subsubsection{Other Exemplar eHealth Applications:} Emotion recognition and stress monitoring are two other widely used eHealth applications that rely on input data similar to that of the pain assessment application. Emotion recognition uses verbal inputs, and cues such as tone of voice, facial expressions, postures, gestures and also through physiological signals \cite{koelstra2011deap}. Stress monitoring application detects the existence of stress in each period of time using physiological signals \cite{han2020objective}. The Electrodermal Activity (EDA) or Galvanic Skin Response (GSR) is one of the physiological signals generated by the human body, which can be used to detect stimuli in individuals. Fall detection is another exemplar application that uses 3-d accelerometer data to detect falls by using classification models \cite{chatzaki2016human}. The input data quality assessment, using multi-modal ML models, configuration of ML models, and edge orchestration techniques used for pain assessment an be analogously applied to other similar applications of emotion recognition, stress monitoring, and fall detection.

\section{Edge-centric Optimization of ML-based eHealth Workloads}\label{sec.edge}


Machine learning (ML) is advancing real-time and interactive user services in healthcare domain~\cite{schmidhuber2015deep}. ML applications are primarily deployed on cloud infrastructure to meet the compute intensity and storage requirements of ML algorithms, and address the resource constraints of user-end wearable sensory devices ~\cite{barbera2013offload}. However, unpredictable network constraints including variable signal strength and availability of the network affect real-time delivery of cloud services~\cite{khelifi2018bringing}. The edge computing paradigm allows deployment of ML applications closer to the user-end devices, minimizing the latency of service delivery, reducing the total network load, and alleviating privacy concerns. 

\textit{Systems Perspective} Collaborative sensor-edge-cloud architecture presents multiple execution choices for workload partitioning and compute placement including execution on a single sensor, edge, cloud nodes, and any possible combinations of these devices. Considering the variable accuracy nature of ML algorithms, these execution choices expose a wide range of energy-performance-accuracy trade-off space. Choosing an optimal execution option under varying system dynamics, available energy budget of devices, network constraints, and error resilience of ML workloads is a complex run-time challenge. 

\textit{Application Perspective} Depending on the ML model employed, different applications feature varying compute, data, and communication intensities. At an application level, there is diversity in terms of sensitivities to latency, throughput, infrastructure availability, and accuracy. Further, different application execution choices result in different energy consumption patterns \cite{naeini2018edge}. Considering these application level variations, the choice of execution of an application is a subject of multiple factors varying at run-time \cite{gia2015fog}. 

\textit{Edge Orchestration} Edge orchestration techniques handle workload partitioning, distribution, and scheduling of ML workloads, considering both applications' and systems' perspectives. Understanding the varying compute intensities of applications, latency and throughput requirements, user-interaction and responsiveness expectations, quality of interconnection network including signal strength, availability, load balancing, compute and storage capacities of underlying hardware elements put together makes application orchestration a stochastic process~\cite{shahhosseini2021towards,shahhosseini2021exploring}. To maximize the efficiency of edge-enabled health care services, run-time solutions that holistically consider both requirements and opportunities vertically across the user, device, application, network, edge node, and platform layers are necessary \cite{seo2020dynamic}. In this Section, we present state-of-the-art edge orchestration techniques for efficient compute placement with rule-based heuristics, and optimized orchestration through reinforcement learning.




\subsection{Dynamics of Compute Placement} 

Computation offloading techniques transfer the execution an application, or a task within an application, to a resourceful device for improving performance \cite{mach2017mobile}. Compute placement determines the choices on partitioning an application, and selection of the external resourceful edge/cloud nodes onto which the partitioned task is to be offloaded \cite{barbera2013offload}. Some of the existing compute placement and offloading strategies do not consider the diversity in applications' compute and communication requirements, eventual performance gain with offloading and compute placement choices, and potential latency penalties incurred with those choices. For optimal performance gains, compute placement techniques have to consider the dynamically varying application and network characteristics, energy budgets, and accuracy-performance tradeoffs \cite{eshratifar2019jointdnn,shahhosseini2019dynamic,teerapittayanon2017distributed}.


We explore the intricacies of running smart eHealth applications on multi-layer sensor-edge-cloud platforms to demonstrate the dynamics of compute placement. We choose stress monitoring \cite{multimodal_stress_embc2019}, fall detection \cite{chatzaki2016human}, and pain assessment \cite{han2020objective} as representative workloads from ML-driven digital healthcare systems. The stress monitoring application uses predictive models to extract statistical features from physiological parameters of Electro-dermal Activity (EDA) and Galvanic Skin Response (GSR), and predict stress levels \cite{aqajari2020gsr}. The fall detection application uses de-noising, feature extraction, and decision-tree training for classification of fall and no-fall events \cite{seo2020dynamic}. The pain assessment application uses pre-processing, feature extraction, and SVM classification for determining level of pain \cite{laitala2020rpeak}. In summary, each application typically includes the pipeline of data pre-processing, feature extraction, and classification ML tasks. We execute these workloads on real hardware testbed that emulates a baseline sensor-edge-cloud platform. We consider an edge platform with configurable on-board sensors for sensing at variable sampling rates, and connectivity to cloud infrastructure. We consider three compute placement policies with the execution choices of running the ML workloads: i) fully on the edge device, ii) fully on the cloud node, iii) partially on the edge and partially on the cloud. For evaluation, we define \textit{latency} metric as the time taken to respond to a request including the communication and execution costs.

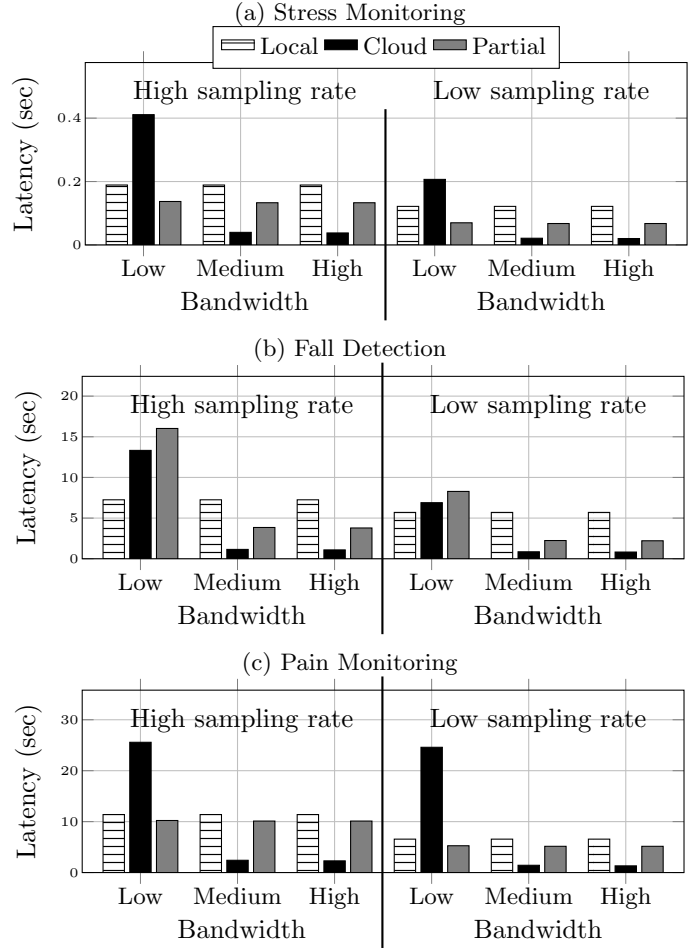
\begin{figure}[ht!]
\pgfplotsset{every x tick label/.append style={font=\tiny, yshift=0.5ex}}
\pgfplotsset{every y tick label/.append style={font=\tiny, xshift=0.5ex}}
\usetikzlibrary{patterns}
\pgfplotsset{compat=1.11,
	/pgfplots/ybar legend/.style={
		/pgfplots/legend image code/.code={%
			\draw[##1,/tikz/.cd,yshift=-1mm]
			(5mm,2mm) rectangle (2pt,0.2em);},
	},
}
\centering
\begin{subfigure}[b]{\textwidth}\caption{Stress Monitoring} \label{fig:Stress}
\centering
\vspace{-2.5mm}
\begin{tikzpicture}
\begin{axis}[
ybar=2pt,
enlarge x limits={abs=0.6},
ymin=0,
width  = 9.5cm,
height = 4cm,
bar width=8pt,
grid=major,
ylabel={\normalsize Latency (sec)},
xticklabel style={rotate=0, font=\small},
xtick = data,
ylabel near ticks,
table/header=false,
every node near coord/.append style={font=\tiny},
table/row sep=\\,
xticklabels from table={
	Low \\
	Medium\\
	 High\\
	Low \\
	Medium\\
	 High\\
}{[index]0},
legend columns=3,
enlarge y limits={value=.4,upper},
legend style={at={(0.5,1.2)},anchor=north, font=\small},
]
\addplot [draw=black,pattern=horizontal lines] table[x expr=\coordindex,y index=0]{0.189038\\0.189038\\0.189038\\0.121628\\0.121628\\0.121628\\};
\addplot [draw=black,fill=black]  table[x expr=\coordindex,y index=0,red]{0.410974\\0.039724\\0.037849\\0.20681\\0.021185\\0.0202475\\};
\addplot [draw=black,fill=gray] table[x expr=\coordindex,y index=0]{0.137094\\0.133134\\0.133114\\0.069862\\0.067585\\0.0675735\\}; 
\legend{Local, Cloud, Partial}
\pgfplotsinvokeforeach{0,1,2,3}{\coordinate(l#1)at(axis cs:#1,0);}
\end{axis}
\draw[thick] (3.95,-1) -- (3.95,1.8);
\node[below] at (2.1,2.3) {High sampling rate};
\node[below] at (6.0,2.3) {Low sampling rate};
\node[below] at (2.1,-0.5) {Bandwidth};
\node[below] at (6.0,-0.5) {Bandwidth};
\end{tikzpicture}
\vspace{1.5mm}
\end{subfigure}

\begin{subfigure}[b]{\textwidth}\caption{Fall Detection} \label{fig:Fall-b}
\centering
\vspace{-2.5mm}
\begin{tikzpicture}
\begin{axis}[
ybar=2pt,
grid=major,
enlarge x limits={abs=0.6},
ymin=0,
width  = 9.5cm,
height = 4cm,
bar width=8pt,
ylabel={\normalsize Latency (sec)},
xticklabel style={rotate=0, font=\small},
xtick = data,
ylabel near ticks,
table/header=false,
every node near coord/.append style={font=\tiny},
table/row sep=\\,
xticklabels from table={
		Low \\
	Medium\\
	 High\\
	Low \\
	Medium\\
	 High\\
}{[index]0},
legend columns=3,
enlarge y limits={value=.4,upper},
legend style={at={(0.5,.99)},anchor=north, font=\small}
]
\addplot [draw=black,pattern=horizontal lines] table[x expr=\coordindex,y index=0]{7.25\\7.25\\7.25\\5.7\\5.7\\5.7\\}; 
\addplot [draw=black,fill=black]  table[x expr=\coordindex,y index=0,red]{13.327\\1.15\\1.0885\\6.895\\0.856\\0.8255\\};
\addplot [draw=black,fill=gray] table[x expr=\coordindex,y index=0]{16.023\\3.846\\3.7845\\8.28\\2.241\\2.2105\\}; 
\pgfplotsinvokeforeach{0,1,2,3}{\coordinate(l#1)at(axis cs:#1,0);}
\end{axis}
\draw[thick] (3.95,-1) -- (3.95,2.6);
\node[below] at (2.1,2.3) {High sampling rate};
\node[below] at (6.0,2.3) {Low sampling rate};
\node[below] at (2.1,-0.5) {Bandwidth};
\node[below] at (6.0,-0.5) {Bandwidth};
\end{tikzpicture}
\vspace{1.5mm}
\end{subfigure}

\begin{subfigure}[b]{\textwidth}\caption{Pain Monitoring} \label{fig:Fall}
\vspace{-2.5mm}
\centering
\begin{tikzpicture}
\begin{axis}[
ybar=2pt,
grid=major,
enlarge x limits={abs=0.6},
ymin=0,
width  = 9.5cm,
height = 4cm,
bar width=8pt,
ylabel={\normalsize Latency (sec)},
xticklabel style={rotate=0, font=\small},
xtick = data,
ylabel near ticks,
table/header=false,
every node near coord/.append style={font=\tiny},
table/row sep=\\,
xticklabels from table={
	Low \\
	Medium\\
	 High\\
	Low \\
	Medium\\
	 High\\
}{[index]0},
legend columns=3,
enlarge y limits={value=.4,upper},
legend style={at={(0.5,.99)},anchor=north, font=\small},
]
\addplot [draw=black,pattern=horizontal lines] table[x expr=\coordindex,y index=0]{11.394\\11.394\\11.394\\6.564\\6.564\\6.564\\}; 
\addplot [draw=black,fill=black]  table[x expr=\coordindex,y index=0,red]{25.58\\2.414\\2.297\\24.598\\1.432\\1.315\\};
\addplot [draw=black,fill=gray] table[x expr=\coordindex,y index=0]{10.222\\10.1131\\10.11255\\5.265\\5.1561\\5.15555\\}; 
\pgfplotsinvokeforeach{0,1,2,3}{\coordinate(l#1)at(axis cs:#1,0);}
\end{axis}
\draw[thick] (3.95,-1) -- (3.95,2.6);
\node[below] at (2.1,2.3) {High sampling rate};
\node[below] at (6.0,2.3) {Low sampling rate};
\node[below] at (2.1,-0.5) {Bandwidth};
\node[below] at (6.0,-0.5) {Bandwidth};
\end{tikzpicture}
\vspace{-1.5mm}
\end{subfigure}
\caption{Latency with different compute placement strategies for eHealth applications under different network bandwidth (low, medium, high) and sampling rates (high and low). (a) Stress monitoring, (b). Fall detection, (c). Pain monitoring.}
\vspace{-5mm}
\label{Exploration}
\end{figure}

Figure \ref{Exploration} shows the latency of stress monitoring, fall detection, and pain monitoring applications with different compute placement choices, over different network bandwidths, and sampling rates of input sensory devices. The compute placement choices Local, Cloud, Partial represent execution of applications on edge node, cloud node, and collaborative edge-cloud nodes. We used three levels of available bandwidths -- low, medium, and high. Available bandwidth influences the latency of execution, specifically incurred in transmitting data from edge to cloud nodes. We used two levels of sampling rates -- high and low, for input data sensory devices. Sampling rate determines the total volume of data being transmitted from edge to cloud nodes, effecting the latency.

The evaluation shows that network variation in terms of available bandwidth substantially effects the latency and choice of compute placement. For example, consider the scenario shown in Figure \ref{Exploration} (a), under high sampling rate and high bandwidth. In this case, compute placement on the cloud is the optimal choice in comparison with the edge, and partial choices. Since there is sufficient network bandwidth, the penalty of transmitting data from edge to cloud is minimal, and the performance gain of executing the application on the cloud is also significant. In contrast, for the same scenario with low bandwidth (LBW), compute placement on the cloud has the highest latency, owing to the higher penalty of transmitting data from edge to cloud under low bandwidth. In this case, the partial (edge-cloud) execution has a better latency. 

In addition to the bandwidth availability, the application's nature can substantially influence the choice of compute placement. For instance, consider the scenario in Figure \ref{Exploration} (b), under high sampling rate and low bandwidth. The fall detection application's compute-communication ratio, and lower bandwidth makes the local execution optimal, as opposed to offloading to the cloud node. 

The sensing configuration of an application alters the volume of data generated for transmission and processing. For instance, consider the scenario in Figure \ref{Exploration} (c), under low sampling rate and medium bandwidth. As the sampling rate is lower, the penalty incurred in transmitting data to the cloud node is minimized. With lower data volume and availability of either medium (and/or high) bandwidth results in significant improvement in latency. 

It should be noted that lowering the sampling rate potentially sacrifices the accuracy, although the latency is improved. While the accuracy loss is subjective to the error resilience of the application, missing insightful input data samples could lead to mis-predictions and critical errors. Mis-predictions can be minimized by setting an upper bound on accuracy requirements and/or bounds on lowering the sampling rates. In such scenarios, the error resilience of an application also influences the compute placement decisions. 


\subsection{Using RL for Optimization}


Finding the optimal orchestration policy for an unknown and dynamic system is critical since dynamicity of environment (e.g., network condition, workload arrival at computing nodes, user traffic, and application characteristics) changes over time. Most current solutions are based on design time optimization, without considerations on varying system dynamics at run-time~\cite{zhang2016energy,cao2018joint,mao2016power,nan2018dynamic,chamola2017latency,chang2017energy,bao2017cost,kattepur2016resource,you2018exploiting,sheng2015energy,liu2016delay}. A complex system that runs a variety of applications in uncertain environmental conditions requires dynamic control to offer high-performance or low-power guarantees~\cite{seo2020dynamic,naeini2018edge,shahhosseini2022online,shahhosseini2022hybrid,shahhosseini2019dynamic}. Considering the run-time variation of system dynamics, and making an optimal orchestration choice requires intelligent monitoring, analysis, and decision making. Existing heuristic and rule-based orchestration methods require an extensive design space exploration to make optimal compute placement decisions. Further, such a solution based on exhaustive search at run-time becomes practically infeasible for latency critical services. In this context, different offline and online machine learning models have been adopted for run-time resource management of distributed systems, to handle the complexity of orchestration choices. Among these models, reinforcement learning approach (RL) is effective in developing an understanding and interpreting varying system dynamics ~\cite{mousavi2016deep,park2019wireless}. Reinforcement learning enables identification of complex dynamics between influential system parameters, and online decision making to optimize objectives such as response time, energy consumption, and quality of service \cite{sutton2018reinforcement}. RL approach allows formulating policies at run-time using the input data collected over time. Specifically, RL approach uses a reward function to quantify the effect of an action on the system state. This allows optimizing orchestration choices over time, considering the system wide context and objectives. In this sub-section, we present the design of reinforcement learning agent for orchestrating ML workloads on sensor-edge-cloud platforms. 


 \begin{figure}[tb!]
    \centering
      \includegraphics[width=\linewidth]{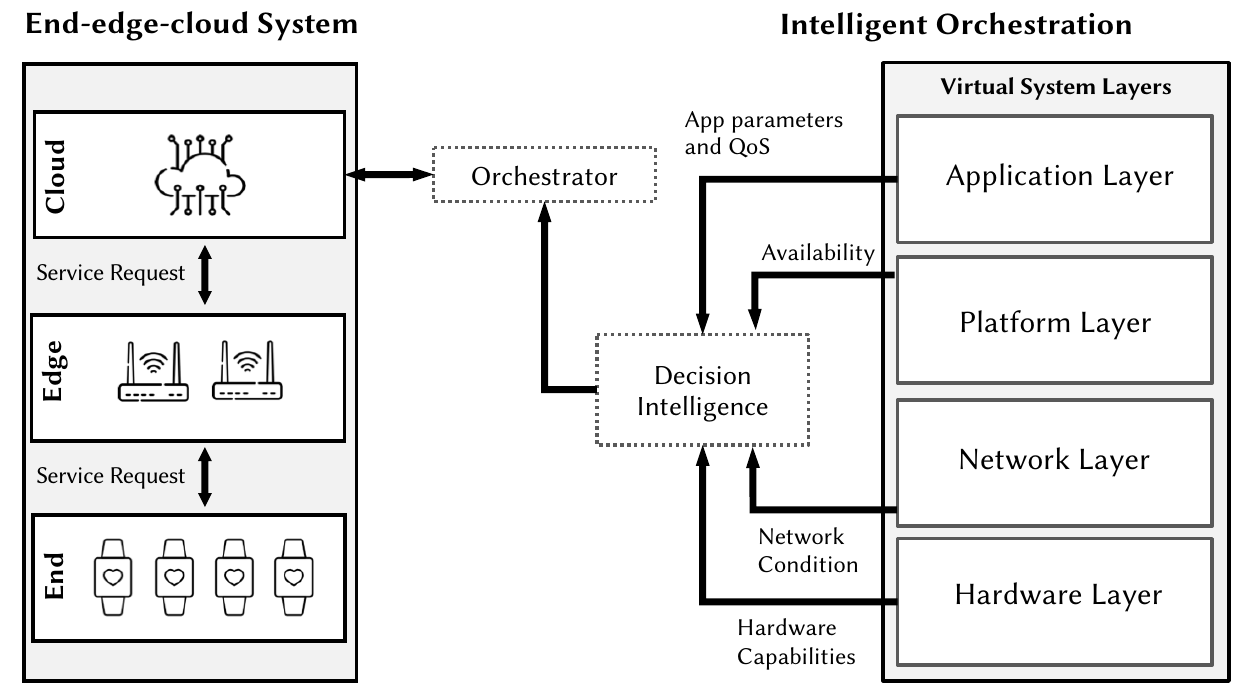}
      \caption{Overview of intelligent orchestration in end-edge-cloud architectures }
      \label{fig.sys_arch}
    \end{figure}
\subsubsection{Orchestration Framework:}
Figure \ref{fig.sys_arch} shows an overview of generic intelligent orchestration framework for multi-layered end-edge-cloud architectures \cite{shahhosseini2022online}. This framework uses system-wide information for intelligent orchestration through virtual system layers that include application, platform, network and hardware layers. Each of the virtual system layers provide inputs for monitoring system and application dynamics such as application adjustment parameters, accuracy requirements, availability of devices for execution, network characteristics, and hardware capabilities. Each execution choice affects the performance and energy consumption of the user end-device, based on the system parameters such as hardware capabilities, network conditions, and workload characteristics. Each layer exhibits a diverse set of requirements, constraints, and opportunities to trade-off performance and efficiency that vary over time. For example, the application layer focuses on the user's perception of algorithmic correctness of services, while the platform layer focuses on improving system parameters such as energy drain and data volume migrated across nodes. Both application and platform layers have different measurable metrics and controllable parameters to expose different opportunities that can be exploited for meeting overall objectives. The network layer provides connectivity for data and control transfer between different physical devices. In addition, the hardware layer provides hardware capabilities for computing nodes in the system. Run-time system dynamics affect orchestration strategies significantly in addition to requirements and opportunities. Sources of run-time variation across the system stack include workload of a specific computing node, connectivity and signal strength of the network, mobility and interaction of a given user, etc. Information on run-time cross-layer requirements and run-time variations provide necessary feedback to make appropriate decisions on system configurations such as offloading policies.
\subsubsection{RL Agent for Orchestration:}
Making the optimal orchestration choice considering these varying dynamics is an NP-hard problem, while brute force search of a large configuration space is impractical for real-time applications. Understanding the requirements at each level of the system stack and translating them into measurable metrics enable appropriate orchestration decision making. On the other hand, heuristic, rule-based, and closed-loop feedback control solutions are slow in convergence due to the large state space~\cite{sutton2018reinforcement}. To address these limitations, reinforcement learning (RL) approaches have been adapted for the computation offloading problem~\cite{sen2019machine}. RL builds specific models based on data collected over initial epochs, and dramatically improves the prediction accuracy~\cite{sutton2018reinforcement}. We design an RL agent that monitors system-wide parameters and chooses a suitable action that maximizes the efficiency of orchestration decisions. 

\begin{figure}[ht!]
\centering
  \includegraphics[width=0.7\linewidth]{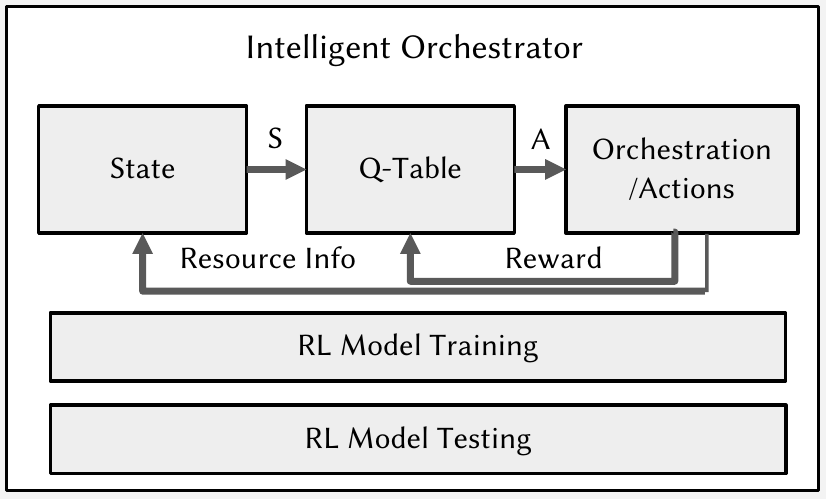}
  \caption{RL agent for intelligent orchestration.}
  \label{fig.rl-agent}
\end{figure}

Figure \ref{fig.rl-agent} shows the workflow of the RL agent for making orchestration decisions. The RL agent component is deployed within the decision intelligence and orchestration blocks in orchestration framework (Figure \ref{fig.3-layer-arch}). The RL agent receives resource information (e.g., processor utilization, available memory, available bandwidth) from the virtual system layers of the orchestration framework (Figure \ref{fig.3-layer-arch}). The RL agent also collects the reward information (response time in this case) from the environment to learn an optimal action that maximizes the reward. The agent builds the Q-Table for Q-Learning algorithm, based on cumulative reward obtained from the environment over time. This signifies the efficacy of a specific orchestration decision (action) in achieving the target of minimizing latency, and enables subsequent optimal orchestration decisions.

We demonstrate the efficacy of using the RL agent for orchestrating ML workloads on sensor-edge-cloud platforms. While the focus of this Chapter is optimizing smart eHealth applications, we use image classification task in the experimentation for the purpose of demonstrating multi-user ML workloads. We implement a scenario where a device-edge-cloud architecture serves up to five end-users to execute ML services  simultaneously. In this scenario, users sends requests for image classification to an agent located at the cloud. In addition, end-users share their resource availability to the agent. The agent decides to orchestrate the ML tasks based on three static (i.e., device-only, edge-only, and cloud-only) and one RL-optimized strategies. In the device-only strategy, each end-device executes the inference service on a local device. Thus, varying number of users has no effect on the average response time in this case. In the edge and cloud only strategies, simultaneous requests compete for edge and cloud resources. This increases the average response time significantly, as the number of users increase. In the \textbf{RL-optimized} strategy, the resource availability is continuously observed and the ML tasks are orchestrated accordingly. 

\begin{figure}[ht!]
\centering
  \includegraphics[width=\linewidth]{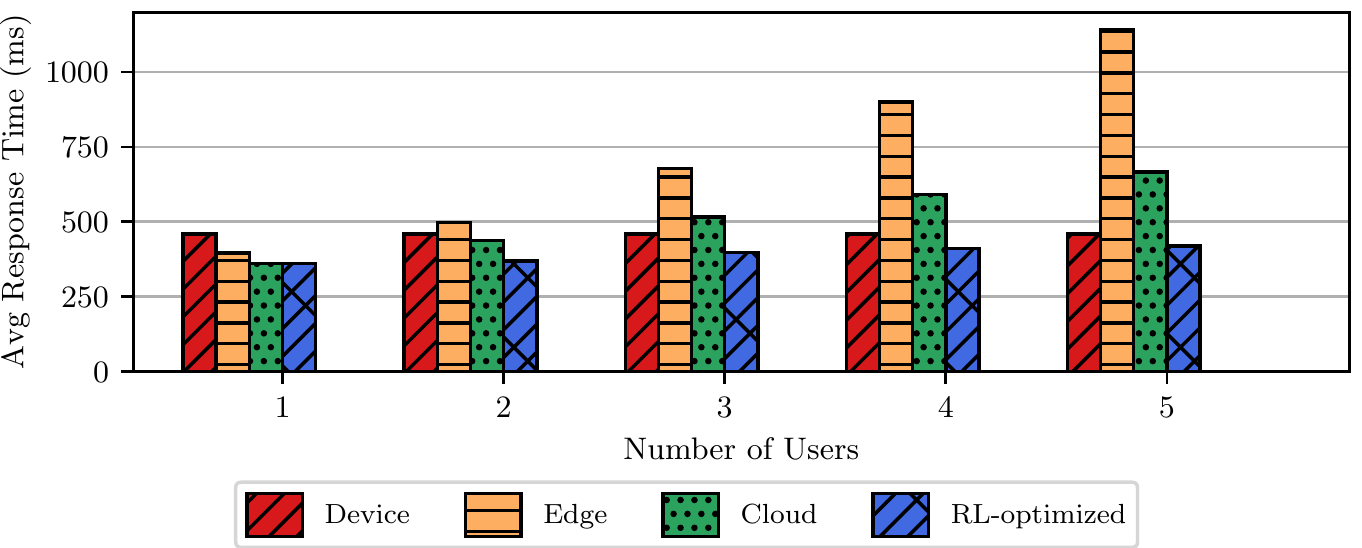}
  \caption{DL inference orchestration in an end-edge-cloud system which runs an image classification application.}
  \label{fig.rloptimized}
\end{figure}

Figure ~\ref{fig.rloptimized} shows the average response time for different numbers of active users for regular network conditions, using different orchestration strategies. The x-axis represents the number of active users. Each bar represents a different orchestration decision made by using the corresponding orchestration strategy~\cite{shahhosseini2022hybrid,shahhosseini2022online}. In RL-optimized approach the average response time remains constant while the number of users is less than three. This is due to the orchestration decision of distributing the services across edge and cloud layers. As the number of users increase to three, the services start competing for resources, leading to an increase in the average response time. With the number of users increasing from three to five, the average response time increases, but at a relatively lower rate, exhibiting efficient utilization of the edge and cloud resources. As the number of users increase, the efficiency of the RL-optimized approach over the static strategies is more prominent.


\section{Sense-Compute Co-optimization of ML-driven eHealth Applications}\label{sec.amser}



\begin{figure}[ht!]
    \centering
    \includegraphics[width=0.98\linewidth]{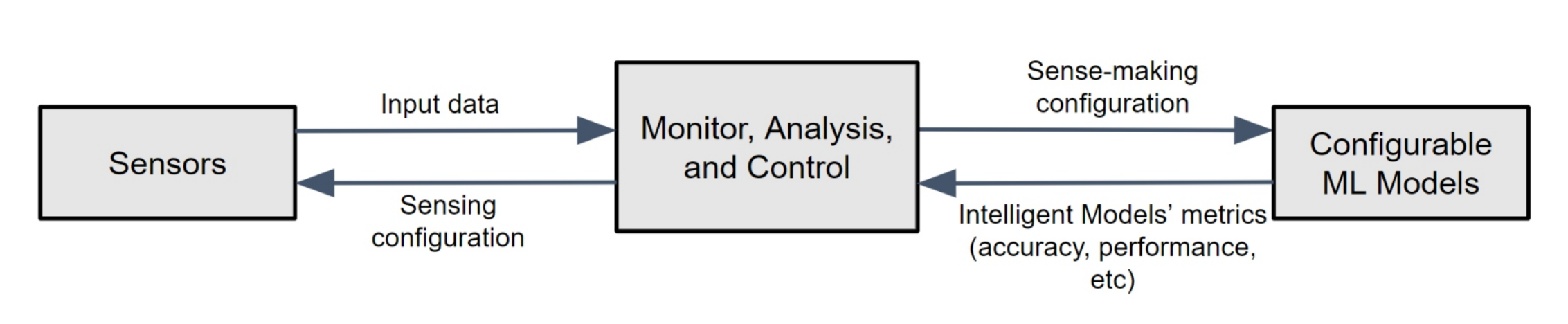}
    \caption{Sensing and Sense-making (compute) modules' interaction with their action knobs in sensor level and intelligent models.}
    \label{fig.simple_sense_compute}
\end{figure}

In this section, we describe sense-compute co-optimization approaches for improving resiliency of smart eHealth applications. Common use cases of ML models often handle complete and clean input data, with no specific sensing challenges. However, using ML methods in smart eHealth applications on edge devices requires considerations on challenges from the sensory data acquisition phase \cite{convergence_edge_ieee}. With different types of implanted, wearable, on-body, and remote sensors, there is a higher probability of noise, motion artifacts, and missing input data from sensors \cite{ma2021smil}. For critical healthcare IoT applications, input data perturbation from motion artifacts, physical failure of sensors, network anomalies and other factors can affect the prediction accuracy of ML models significantly \cite{naeini2019real}). On the other hand, the sense-making (computation) phase of eHealth applications faces the challenge of limited computational capacity of the edge devices for running ML models \cite{convergence_edge_ieee}. Additionally, the ML models should be resilient to probable sensing anomalies like noisy or missing input data, and maintain higher prediction accuracy even with potential garbage input signals \cite{ma2021smil}. Moreover, some applications (e.g., pain assessment in clinical healthcare monitoring systems) require near real-time response time, emphasizing the need for ML inference performance \cite{naeini2018edge}. Addressing these multitude of challenges necessitates a co-optimization approach that jointly handles sensing and sense-making phases for system-wide exploration of suitable optimizations. A simple schematic for the interaction between sensing and sense-making modules is shown in Figure \ref{fig.simple_sense_compute}. Sensing-awareness can be developed by monitoring, and analyzing continuous streams of input data. This intelligence can be used to control the sensing and sense-making configurations simultaneously, by fine-tuning ML models to fit input data characteristics. 

\subsection{Handling Input Data Perturbations}

\subsubsection{Sensing Phase Knobs:} Monitoring input data originating from the sensory devices for anomalies, discrepancies, and noisy components etc., provides insights into quality of input data \cite{sampling_edge_sensors2020}. Understanding the quality of input data can be exploited for selective sensing, such that unreliable sensors can be downsampled, to dampen the effect of unreliable input data. When the input data from a specific sensor is noisy, the sampling rate of the sensory device can be decreased so that the garbage data would not waste the computational resources in the edge layer. In acute scenarios, input data from a specific device might be completely un-reliable (e.g., when a heartbeat sensor is detached from the human body). In this case, detection of the sensor detachment can guide the computation phase to ignore the input from that specific device, and rely on the data from other input modalities \cite{multimodal_stress_embc2019}. It should be noted that both selective sensing and disabling a sensor modality minimizes the network latency penalty with reduced input data volume. In specific scenarios of unreliable network connection, sensing phase knobs can be opportunistically triggered to complement the network delay with reduced input data volume \cite{yick2008wireless}.
\subsubsection{Sense-making Phase Knobs:} Addressing input data perturbation from multiple modalities requires appropriate and proportional actions in the ML algorithmic phase. For instance, selective or greedy feature selection from the pre-processed input data can reduce the noisy data components fed into the ML models \cite{greedy_feature_selection}. This approach is effective in reducing un-necessary computation over noisy input data, although potentially affecting the accuracy of the learning models \cite{naeini2022amser}. Selecting an appropriate ML model from a pool of pre-trained models is another strategy for handling noisy and missing input data. This requires implementation of a model pool, consisting of different ML models that are trained to handle specific combinations of input modalities \cite{naeini2022amser}. Algorithmic optimizations such as meta learning \cite {model_agnostic_fastadapt_ICML2017}, fast reinforcement learning \cite{fastRL_barreto}, and few-shot learning \cite{fewshot_learning} can also be used for providing efficient ML optimizations particularly for edge devices.
\subsubsection{Co-optimization Knobs:}
Besides the presented methods for optimizing sensing and sense-making phases in the edge devices, it is quite important to use co-optimization techniques so that each of these phases complements the other to form a  holistic system with efficiency and robustness. To achieve this goal, meaningful interaction between these phases is required. For example, the sensing information from input data can be used as a trigger to adjust specific sense-making knobs so that the computation models can adapt to the recent changes in the input sources. One example of this approach is using the early-exit technique \cite{EE_branchynet}\cite{earlyexit_bandit} in the neural network when the input data has good quality with negligible noise. In this way, an acceptable confidence threshold can be achieved in less time by skipping deeper layers in neural network models. Moreover, the sensing module in the edge layer can send information about input sources with high or low reliability of their data to the sense-making module, and then the machine learning models in the edge layer can adjust importance weights to those input sources by using attention mechanisms inside their architecture \cite{attention_for_selective_sensing}\cite{vaswani2017attention}. 


\subsubsection{Example Scenarios:}
\begin{figure}[ht!]
    \centering
    \includegraphics[width=0.98\linewidth]{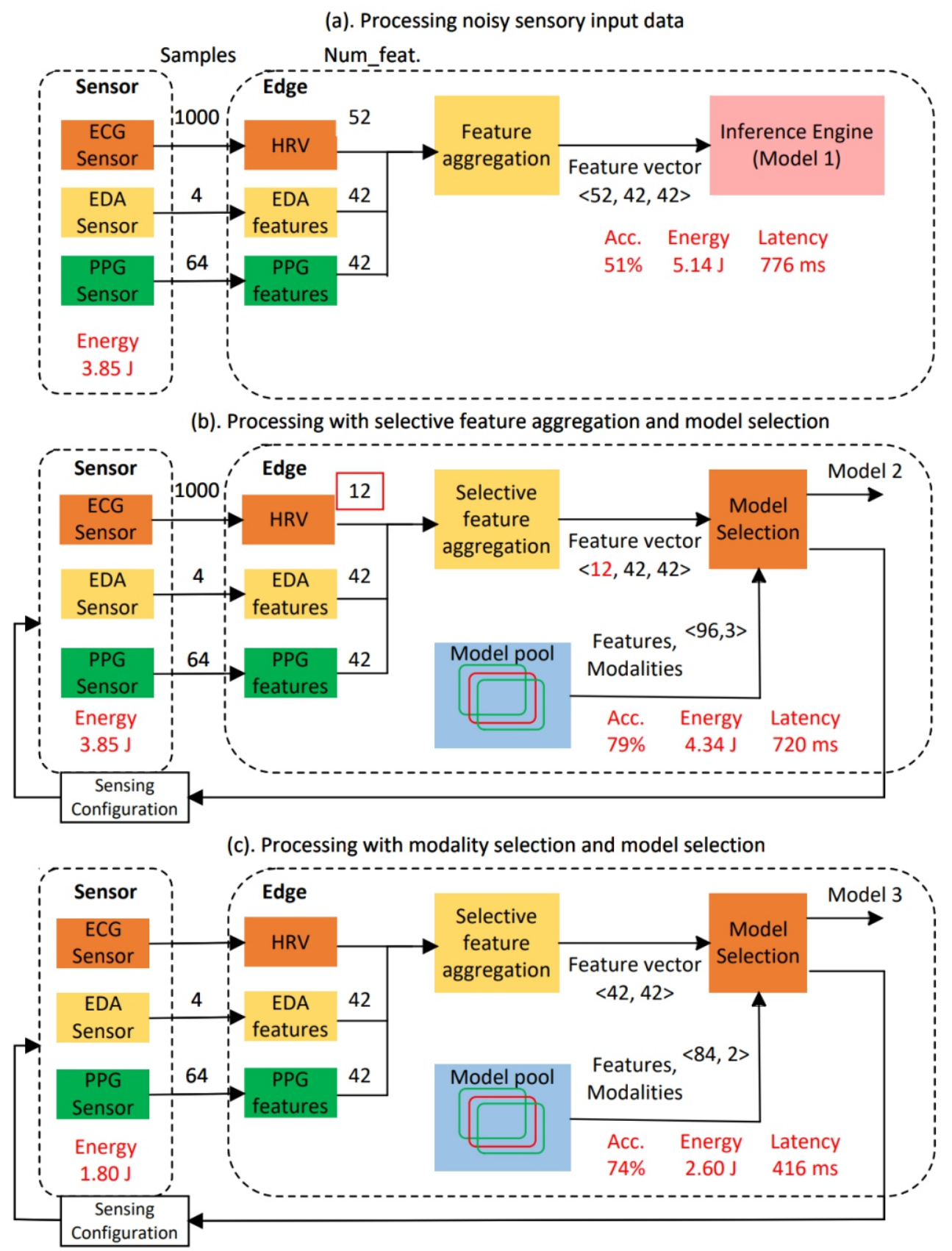}
   \caption{Motivational scenarios for sense-compute co-optimization}
   \label{fig.amser_motiv}
\end{figure}

We demonstrate the advantages of sense-compute co-optimization through an example of multi-modal pain assessment application \cite{naeini2022amser}. The pain assessment application uses inputs from three modalities -- Electrocardiography (ECG), Electrodermal Activity (EDA), and Photoplethysmography (PPG). The ECG, EDA, and PPG modalities have sampling rates of 100, 4, and 64, and generate 52, 42, and 42 features. Figure \ref{fig.amser_motiv} outlines sensor data acquisition, feature extraction, feature aggregation, and inference task for predicting pain levels under different scenarios. Figure \ref{fig.amser_motiv} (a) shows the scenario, where application is executed without modality-awareness. In this scenario, data from the ECG sensor is noisy, yet feature vectors from the noisy modality are fed into the ML model, yielding a baseline prediction accuracy of 51\%. Figure \ref{fig.amser_motiv} (b) shows the scenario where modality-awareness is considered while executing the application. In this scenario, the application is executed with selective feature aggregation, by selecting fewer features from the noisy ECG modality. This reduces the total number of features from the ECG modality to 12. An appropriate ML model to suit the updated feature vector is selected from a model pool, which comprises of pre-trained models. Minimizing the features from noisy ECG modality improves the prediction accuracy to 79\%, while also reducing the energy consumption and improving the performance, in comparison with the baseline scenario (a). Figure \ref{fig.amser_motiv} (c) shows the scenario where modality-awareness is used to select specific modalities with quality input data. In this scenario, the noisy ECG modality is completely dropped, and data from the EDA and PPG modalities is processed. Similar to the scenario (b), an appropriate ML model that suits EDA and PPG inputs is selected from the model pool. Dropping an entire modality of data significantly reduces the computational effort and energy consumption, in comparison with scenarios (a) and (b). It should be noted that the prediction accuracy with only two modalities is 74\%, which is higher than the baseline from scenario (a), while being marginally lower than the prediction accuracy from selective feature aggregation from scenario (b).

\subsection{Sense-Compute Co-optimization Framework}
\begin{figure}[ht!]
    \centering
    \includegraphics[width=0.9\linewidth]{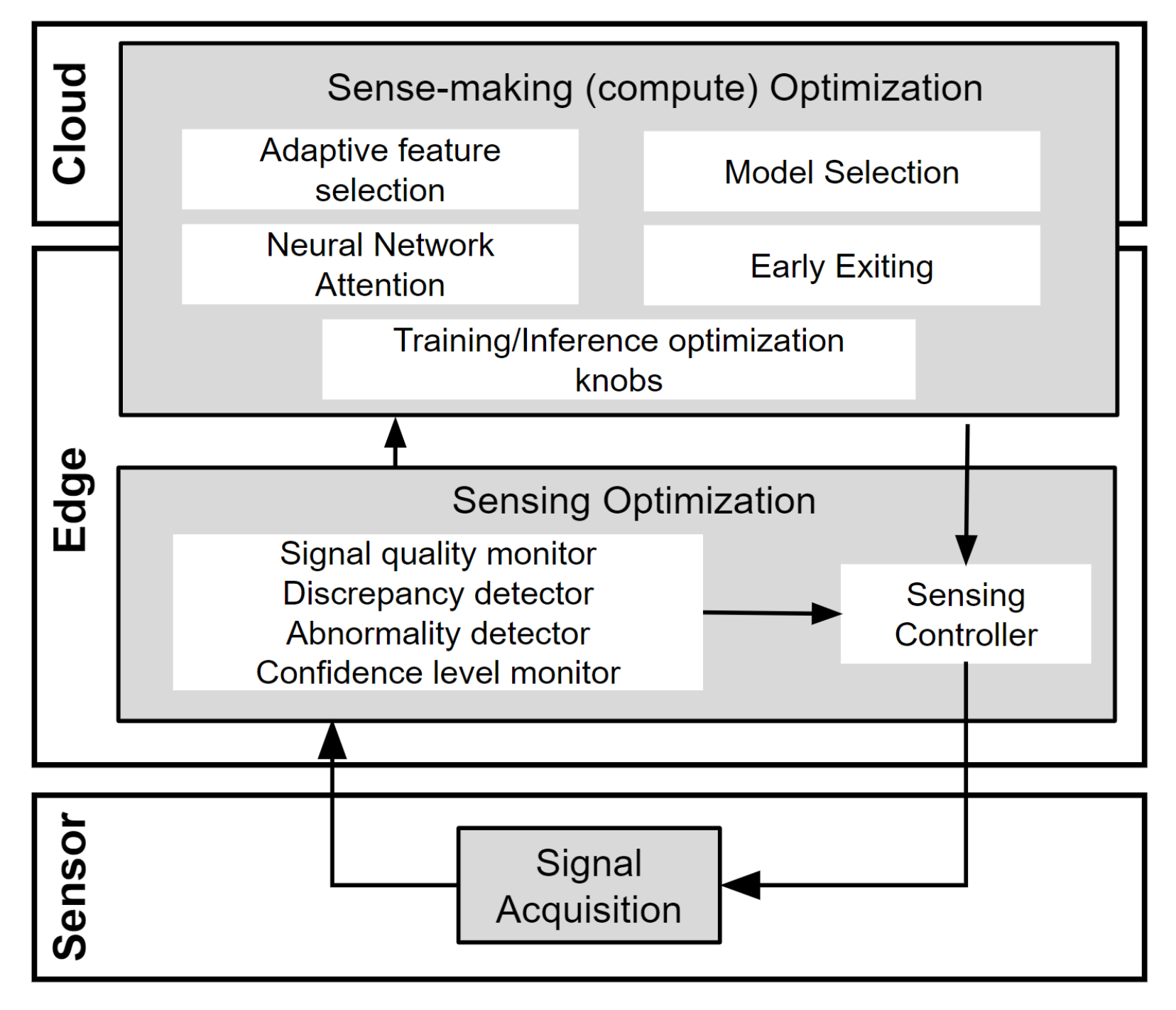}
    \caption{An overview of sensing and sense-making (compute) optimization in a sensor-edge-cloud architecture}
    \label{fig.sense_compute_optimizer}
\end{figure}

 We present the conceptual design of sense-compute co-optimization for multi-modal eHealth applications through the AMSER framework \cite{naeini2022amser}. Figure \ref{fig.sense_compute_optimizer} shows an overview of the AMSER framework for sense-compute co-optimization in multi-layered sensor-edge-cloud platforms. The AMSER framework uses run-time monitoring functionalities of signal quality monitoring, discrepancy detection, abnormality detection, and confidence level monitor to understand the quality of input data modalities, and confidence of the ML model in predicting results. Insights from the run-time monitoring are used to configure the sampling rates of different sensory devices through the sensing controller. Based on the run-time monitoring, the sense-making (compute) optimizer uses ML configuring knobs -- adaptive feature selection, model selection, neural network attention, and early exit mechanisms to configure ML models for current input data sets. 
 For example, data from a specific modality is labeled as uncertain when the signal quality is below a specific Signal to Noise Ratio (SNR). Under such scenarios, the edge level sensing optimizer feeds the learning models with reliable input modalities, while dropping the noisy modality. The sense-making optimizer then selects an ML model from the model pool, that is suitable for the available input data modalities. The model pool contains different pre-trained ML models suitable for different combinations of reliable input modalities. 
 

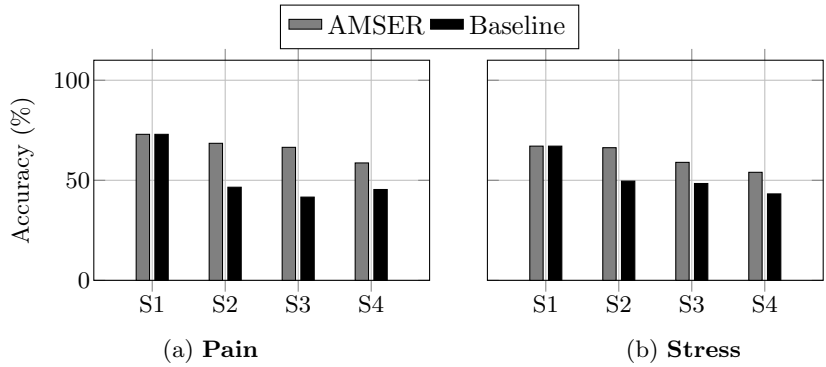
\begin{figure}[ht]
\centering
\pgfplotsset{every x tick label/.append style={font=\tiny, yshift=0.5ex}}
\pgfplotsset{every y tick label/.append style={font=\tiny, xshift=0.5ex}}
\usetikzlibrary{patterns}
\pgfplotsset{compat=1.11,
	/pgfplots/ybar legend/.style={
		/pgfplots/legend image code/.code={%
			\draw[##1,/tikz/.cd,yshift=-1mm]
			(5mm,2mm) rectangle (2pt,0.2em);},
	},
}
\begin{subfigure}[t]{0.45\textwidth}
\begin{tikzpicture}
\begin{axis}[
ybar=2pt,
grid=major,
enlarge x limits={abs=0.8},
ymin=0,
ymax=100,
height = 4.5cm,
bar width=5pt,
ylabel={\small Accuracy (\%)},
xticklabel style={rotate=0, font=\small},
yticklabel style={rotate=0, font=\small},
xtick = data,
ylabel near ticks,
table/header=false,
every node near coord/.append style={font=\tiny},
table/row sep=\\,
xticklabels from table={
	S1\\
	S2\\
	S3\\
	S4\\
}{[index]0},
legend columns=2,
enlarge y limits={value=.1,upper},
legend style={at={(1,1.05)},anchor=south, font=\normalsize},
]
\legend{AMSER, Baseline}
\addplot [draw=black,draw=black,fill=gray] table[x expr=\coordindex,y index=0]{72.97\\68.48\\66.47\\58.67\\}; 
\addplot [draw=black,draw=black,fill=black] table[x expr=\coordindex,y index=0]{72.97\\46.54\\41.6\\45.4\\}; 

\pgfplotsinvokeforeach{0,1,2,3}{\coordinate(l#1)at(axis cs:#1,0);}
\end{axis}
\end{tikzpicture}
\caption{\textbf{Pain}}
\end{subfigure}
\hspace{5mm}
\begin{subfigure}[t]{0.45\textwidth}
\begin{tikzpicture}
\begin{axis}[
ybar=2pt,
grid=major,
enlarge x limits={abs=0.8},
ymin=0,
ymax=100,
height = 4.5cm,
bar width=5pt,
xticklabel style={rotate=0, font=\small},
yticklabels={,,},
xtick = data,
ylabel near ticks,
table/header=false,
every node near coord/.append style={font=\tiny},
table/row sep=\\,
xticklabels from table={
	S1\\
	S2\\
	S3\\
	S4\\
}{[index]0},
legend columns=4,
enlarge y limits={value=.1,upper},
]
\addplot [draw=black,draw=black,fill=gray] table[x expr=\coordindex,y index=0]{67.07\\66.29\\58.96\\53.99\\}; 
\addplot [draw=black,draw=black,fill=black] table[x expr=\coordindex,y index=0]{67.07\\49.6\\48.4\\43.2\\}; 

\pgfplotsinvokeforeach{0,1,2,3}{\coordinate(l#1)at(axis cs:#1,0);}
\end{axis}
\end{tikzpicture}
\caption{\textbf{Stress}}
\end{subfigure}
\caption{Accuracy analysis in comparison with baseline \cite{baseline_for_amser} \cite{naeini2022amser})}
\vspace{-2mm}
\label{fig:acc_amser}
\end{figure}

\begin{figure}[ht]
\centering
\pgfplotsset{every x tick label/.append style={font=\tiny, yshift=0.5ex}}
\pgfplotsset{every y tick label/.append style={font=\tiny, xshift=0.5ex}}
\usetikzlibrary{patterns}
\pgfplotsset{compat=1.11,
	/pgfplots/ybar legend/.style={
		/pgfplots/legend image code/.code={%
			\draw[##1,/tikz/.cd,yshift=-1mm]
			(5mm,2mm) rectangle (2pt,0.2em);},
	},
}
\begin{subfigure}[t]{0.45\textwidth}
\begin{tikzpicture}
\begin{axis}[
ybar=2pt,
grid=major,
enlarge x limits={abs=0.8},
ymin=0,
height = 4.5cm,
bar width=5pt,
ylabel={\small Speedup (Baseline=1)},
xticklabel style={rotate=0, font=\small},
yticklabel style={rotate=0, font=\small},
xtick = data,
ylabel near ticks,
table/header=false,
every node near coord/.append style={font=\tiny},
table/row sep=\\,
xticklabels from table={
	S1\\
	S2\\
	S3\\
	S4\\
}{[index]0},
legend columns=2,
enlarge y limits={value=.1,upper},
legend style={at={(1,1.05)},anchor=south, font=\normalsize},
]
\legend{Pain, Stress}
\addplot [draw=black,pattern=horizontal lines] table[x expr=\coordindex,y index=0]{1\\1.10\\1.27\\2.2\\}; 
\addplot [draw=black,fill=gray] table[x expr=\coordindex,y index=0]{1\\1.12\\1.38\\1.64\\}; 

\pgfplotsinvokeforeach{0,1,2,3}{\coordinate(l#1)at(axis cs:#1,0);}
\end{axis}
\end{tikzpicture}
\caption{Performance gain}
\end{subfigure}
\begin{subfigure}[t]{0.45\textwidth}
\begin{tikzpicture}
\begin{axis}[
ybar=2pt,
grid=major,
enlarge x limits={abs=0.8},
ymin=0,
height = 4.5cm,
bar width=5pt,
xticklabel style={rotate=0, font=\small},
yticklabel style={rotate=0, font=\small},
ylabel={\small Data Reduction },
xtick = data,
ylabel near ticks,
table/header=false,
every node near coord/.append style={font=\tiny},
table/row sep=\\,
xticklabels from table={
	S1\\
	S2\\
	S3\\
	S4\\
}{[index]0},
legend columns=4,
enlarge y limits={value=.1,upper},
]
\addplot [draw=black,draw=black,pattern=horizontal lines] table[x expr=\coordindex,y index=0]{1\\1.56\\2.79\\3.19\\}; 
\addplot [draw=black,draw=black,fill=gray] table[x expr=\coordindex,y index=0]{1\\1.33\\3.17\\7.12\\}; 

\pgfplotsinvokeforeach{0,1,2,3}{\coordinate(l#1)at(axis cs:#1,0);}
\end{axis}
\end{tikzpicture}
\caption{Data volume reduction}
\end{subfigure}
\vspace{-2mm}
\caption{(a) Performance analysis of the edge-layer device for AMSER framework vs. Baseline study. (b) Data transfer volume between the sensors and edge-layer device for AMSER framework vs. Baseline study.}
\vspace{-6mm}
\label{fig.perfdata}
\end{figure}
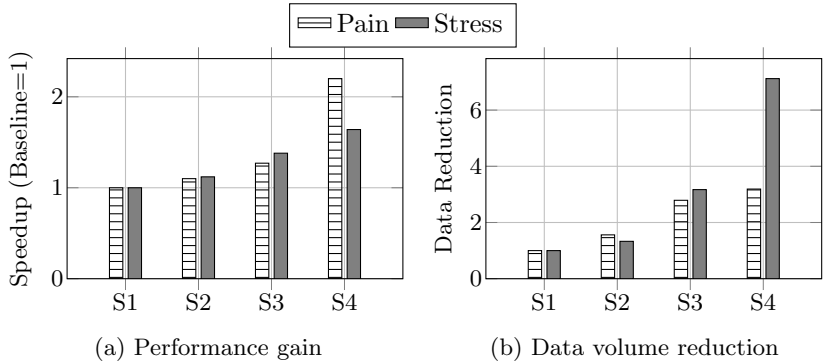

The results for accuracy and performance (speedup) gain with sense-compute co-optimization in comparison with the baseline \cite{baseline_for_amser} is shown in Figures \ref{fig:acc_amser} and \ref{fig.perfdata}. ML-driven eHealth applications of pain assessment and stress monitoring applications were used as input workloads. For a comprehensive evaluation, four different scenarios (S1-S4) with different noise components in input data of modalities are used. Scenario 1 (S1) is the baseline with no noise components. Scenario 2 (S2), has a wandering noise added to the original data. Scenario 3 has an additional motion artifact added to one input modality on top of the wandering noise, which makes that modality completely unreliable. In scenario 4 (S4), two modalities suffer from severe motion artifact noise. In scenario 2 (S2), the AMSER framework activates the selective feature selection to handle the noisy input modalities, resulting in higher prediction accuracy in comparison with the baseline. In scenario 3 with unreliable input (S3), the AMSER approach drops the entire unreliable modality, yielding a better prediction accuracy and performance gain. Similarly, the AMSER framework drops two noisy modalities in scenario 4 (S4), resulting in better prediction accuracy, and significantly higgher performance. The sensing awareness and synergistic compute knob actuation of the AMSER platform provides significant improvements in accuracy, efficiency, and performance, in comparison with existing dis-joint sensing and compute optimization approaches. 


\section{Conclusions} \label{sec.conclusion}
In this chapter, we  presented edge-centric optimizations for ML-driven eHealth applications through compute placement, improving the compute placement decisions through reinforcement learning agent, and cross-layered sense-compute co-optimizations. We also presented an exemplar case study of objective pain assessment to demonstrate the use cases of edge-ML based smart eHealth applications, common data flows, computation challenges, and frameworks for deploying smart eHealth applications.

\subsection{Key Insights}
ML-driven smart healthcare applications have different input data characteristics, computational requirements, and quality metrics. Continuous stream of input data, varying network conditions, and computational requirements of different ML models create dynamic workload scenarios. At an application-level, requirements include higher prediction accuracy of ML models, latency of inferencing results from ML models, resilience, and an overall higher quality of service. At a system-level, requirements include availability of compute nodes in edge and cloud layers, compute capabilities of edge nodes to meet performance requirements of ML models, network utilization, and overall energy efficiency. Considering both application and system-level parameters simultaneously is necessary for optimizing edge-centric ML-driven smart eHealth applications. Optimized compute placement has higher efficacy in meeting both application and system requirements simultaneously. Accuracy-performance trade-offs can also be explored within the compute placement phase, by configuring the choice of ML models. Both model-based and model-free reinforcement learning agents can guide the compute placement decisions on choice of execution node and tuning accuracy-performance trade-offs with high degree of convergence. Further, multi-modal eHealth applications are prone to input data perturbations, which also presents an opportunity to exploit the inherent resilience to selectively process input data. This brings sensing-awareness into computation, and compute-awareness to sensing through bi-directional feedback. Cross-layered sense-compute co-optimization improves sensing, computation, and communication aspects of edge-ML based eHealth applications holistically. 

\subsection{Open Research Directions}
\subsubsection{Data Quality Management:} Input data quality is an essential component for improving prediction accuracy of ML-driven smart eHealth applications. Processing exclusively quality input data also improves the bandwidth utilization and latency of tasks run on the edge nodes. Some of the techniques presented in this chapter address the sensing aspects through continuous monitoring and analysis of input data quality. Qualitative assessment of sensory data can be improved significantly beyond the rule-based monitors using cognitive learning models. Design of autonomous models for input data quality management remains an open challenge. Autonomous models enable reasoning for different input perturbations to assess true quality of sensory inputs. Consequently, the garbage data that is un-necessarily processed is minimized, supporting the scalability of edge ML solutions. Quality assessment of input data is significant in other sensor-driven domains such as autonomous driving, robotics, and computer vision etc.
\subsubsection{Contextual Edge Orchestration:} Orchestration techniques for collaborative sensor-edge-cloud architectures improve a multitude of metrics in terms of performance, turn around time, energy efficiency, accuracy trade-off exploration, and network utilization. Orchestration techniques presented in this chapter enable intelligent compute placement, off-loading, and accuracy configuration decisions through rule-based heuristics. However, contextualization of system dynamics to reason for orchestration decisions, and exploration of accuracy-performance-energy trade-offs with context-awareness is an open research direction. Reinforcement learning has been widely used for optimal orchestration decisions at run-time, considering the varying system dynamics. The efforts in collecting training data, online updating, and convergence time influence the efficacy of such learning methods. In this perspective, design of model-free, and few-shot learning models for run-time edge orchestration is a promising open research direction. 
\subsubsection{Sense-Compute Co-optimization:} Cross-layered sense-compute co-optimization is the most effective strategy for improving sensor dependant, edge-based ML applications. In this chapter, we presented adaptive sensing and sensing-aware computing techniques that uses system-wide monitoring and intelligence for sense-compute co-optimization. This approach focuses on selecting appropriate ML models based on quality of input modalities, exploiting the inherent resilience of multi-modal ML applications. Adaptive feature selection, and ML model selection enforce the idea of sense-compute co-optimization at a coarse-grained level, and require multiple pre-trained models. Incorporating fine-grained edge-layer ML model configurations such as early exit, greedy feature selection, neural network model attention, and saliency maps etc., can complement the model selection strategy. The feasibility of such edge-layer based ML model tuning in collaboration with cloud-layer based model selection is another open research direction.

\bibliographystyle{splncs03}

\bibliography{references}

\end{document}